\documentclass[10pt,twocolumn,letterpaper]{article}

\usepackage{cvpr}
\usepackage{times}
\usepackage{epsfig}
\usepackage{graphicx}
\usepackage{amsmath}
\usepackage{amssymb}
\usepackage{subcaption}

\usepackage{url}
\usepackage{relsize}
\usepackage{xcolor,colortbl}
\usepackage{multirow}
\usepackage{wrapfig}
\usepackage{bbm}
\usepackage[labelfont=bf]{caption}
\usepackage{tabularx}
\usepackage{kotex}

\usepackage[pagebackref=true,breaklinks=true,letterpaper=true,colorlinks,bookmarks=false]{hyperref}

\cvprfinalcopy 

\ifcvprfinal\fi
\begin{document}

\title{\vspace{-1cm}Developing a Compressed Object Detection Model based on YOLOv4 for\\
 Deployment on Embedded GPU Platform of Autonomous System \vspace{-.25cm}}

\author{Issac Sim$^{1)}$, Ju-Hyung Lim$^{1)}$, Young-Wan Jang$^{1)}$, JiHwan You$^{2)}$, SeonTaek Oh$^{2)}$, Young-Keun Kim$^{^{*}1)}$\\
\small{1) School of Mechanical and Control Engineering, Handong Global University, Pohang 558, Korea}\\ 
\small{2) Department of Mechanical and Control Engineering, Handong Global University, Pohang 558, Korea}
}

\maketitle

\begin{abstract}
\vspace{-.25cm}
Latest CNN-based object detection models are quite accurate but require a high-performance GPU to run in real-time. They still are heavy in terms of memory size and speed for an embedded system with limited memory space. Since the object detection for autonomous system is run on an embedded processor, it is preferable to compress the detection network as light as possible while preserving the detection accuracy. There are several popular lightweight detection models but their accuracy is too low for safe driving applications.  Therefore, this paper proposes a new object detection model, referred as YOffleNet, which is compressed at a high ratio while minimizing the accuracy loss for real-time and safe driving application on an autonomous system. The backbone network architecture is based on YOLOv4, but we could compress the network greatly by replacing the high-calculation-load CSP DenseNet with the lighter modules of ShuffleNet. Experiments with KITTI dataset showed that the proposed YOffleNet is compressed by 4.7 times than the YOLOv4-s that could achieve as fast as 46 FPS on an embedded GPU system(NVIDIA Jetson AGX Xavier). Compared to the high compression ratio, the accuracy is reduced slightly to 85.8\% mAP, that is only 2.6\% lower than YOLOv4-s. Thus, the proposed network showed a high potential to be deployed on the embedded system of the autonomous system for the real-time and accurate object detection applications. 
\end{abstract}

\section{서 론}
\vspace{.25cm}

자율주행자동차는 운전자의 편의를 증대할 수 있는 미래 교통수단으로 주목받고 있으며 미래 산업 시장에서 큰 파급 효과를 줄 것으로 예상된다. \cite{lee2016rnrsodhl, jeong2017wkdbfwngod} 자율주행 실현을 위한 핵심 기술은 크게 인지, 판단, 제어 세 부분으로 나뉜다. 이 중에서 최근 딥러닝을 활용한 객체탐지 기술의 발전으로 인해 주행 중 사물을 인식하는 인지 분야에서 큰 도약이 이뤄지고 있다.

영상 데이터를 통한 딥러닝 기반 객체탐지 알고리즘은 사람의 시각을 통한 검출보다 더 높은 인식률을 보이고 있다. 특별히, 영상 데이터의 경우 라이다 센서에 비해 저렴한 카메라 센서를 통해 데이터를 획득 가능하기 때문에 자율주행자동차의 인지 분야에서 딥러닝을 활용한 객체탐지 알고리즘의 중요도는 점점 증가하고 있다.

딥러닝 기반 객체탐지 알고리즘은 높은 연산 복잡도가 요구되기에 차량의 센서에서 얻은 데이터를 클라우드 서버에 전송하여 처리하는 방식에 의존하고 있다. 그러나 클라우드 컴퓨팅 방식은 해킹에 취약하고 통신망이 구축되지 않은 지역에서는 자율주행이 불가능하다는 한계점을 지닌다. \cite{ferdowsi2018robust} 이러한 문제점을 개선하기 위해 임베디드 환경에서 자율주행에 필요한 연산을 처리하는 엣지(Edge) 컴퓨팅 기술이 요구되고 있다. \cite{liu2019edge} 임베디드 플랫폼에서 딥러닝 기반의 객체탐지 알고리즘이 실시간으로 구동되기 위해서는 한정된 자원에서 동작할 수 있는 경량화된 모델 사이즈를 보유해야 한다. 대표적인 딥러닝 기반 객체탐지 모델인 YOLO, SSD의 경우에도 임베디드 플랫폼에 적용하기에는 모델의 크기가 크고 추론 속도가 느린 한계를 지니고 있다. 이러한 문제를 해결하기 위해서는 정확도 손실을 최소화하면서, 임베디드 시스템에서도 동작 가능한 정도로 모델의 크기를 줄일 수 있는 딥러닝 모델 경량화에 관한 연구가 필수적이다. \cite{chlwogns2019thgud, rladudwns2020wkdbfwngod, lee2017study, dhtjsxor2020emfhsdml, oh2020frdet, cheng2017survey, lee2019recent, howard2019searching}

본 연구에서는 자율주행용 임베디드 플랫폼에 탑재 가능한 딥러닝 객체탐지 모델인 YOffleNet을 제안한다. YOffleNet은 대표적인 객체탐지 모델인 YOLOv4의 아키텍처를 기반으로 한다. Backbone Network에서는 CSP DenseNet으로 구성된 CSPDarkent53을 ShuffleNetv2 모듈로 변경하여 파라미터 수를 감소시켰으며, 정보흐름에 이점을 주는 ShuffleNetv2 모듈의 특성을 고려해 Neck 부분의 PANet 구조를 간소화하는 경량화를 진행했다. Head 부분에서는 YOLOv4의 단일단계 검출기 방식을 활용했다.

본 연구에서 제안하는 객체탐지 모델인 YOffleNet의 성능 측정을 위해 대표적인 자율주행 데이터셋인 KITTI 데이터셋을 통해 차량, 보행자, 사이클리스트 클래스에 대한 학습 및 평가를 진행했다. 또한, 임베디드 플랫폼에서의 탑재 적합성을 확인하기 위해 NVIDIA Jetson AGX Xavier 보드와 NVIDIA Jetson Nano 보드에 학습한 모델을 탑재하여 성능 검증을 진행했다. 실험 결과 제안하는 YOffleNet 모델은 85.8의 mAP와 1.9M의 파라미터 수를 보유하였으며, YOLOv4-s 대비 2.6 낮은 mAP를 가졌으나 파라미터 수는 4.8배 적은 것을 확인했다. 또한, Nvidia AGX Xavier보드에서 46 FPS로 구동되는 것을 확인했기에 한정된 자원을 지닌 자율주행용 임베디드 플랫폼 환경에서의 실시간 구현이 가능함을 검증했다.

\vspace{.25cm}
\section{관련 연구}
\vspace{.25cm}

\subsection{CNN 기반 객체탐지 알고리즘}
딥러닝 기반 객체탐지 알고리즘은 크게 단일단계 방식의 검출기(single stage detector)과 이 단계 방식의 검출기(two stage detector)로 나뉜다. 이 단계 방식의 검출기는 영역 제안(Region Proposal) 네트워크를 통해 객체로 추정되는 물체의 위치를 선정한 후, 선정된 영역들에 대해서 객체의 클래스 분류를 수행한다. 대표적인 이 단계 방식의 검출기로는 Faster-RCNN \cite{ren2015faster}, Mask R-CNN \cite{he2017mask} 등이 있으며, 이 단계 방식의 검출기는 많은 연산량으로 인해 실시간으로 사용되기에 어려움이 많다.

반면, 단일단계 방식의 검출기는 객체 분류와 위치 추정을 한 번에 수행하므로 연산량이 적고 빠른 추론이 가능한 장점이 있다. 대표적인 단일 단계 방식의 객체탐지 모델로는 SSD \cite{liu2016ssd}와 YOLO 계열(YOLOv1~v4 \cite{redmon2016you, redmon2017yolo9000, redmon2018yolov3, bochkovskiy2020yolov4})의 모델들이 있다. SSD의 경우 다양한 크기의 convolution layer에 대해 multi-scale feature map을 형성하여 객체탐지를 진행한다. DSSD \cite{fu2017dssd}는 SSD 모델에 deconvolution layer를 추가하여, 속도 성능을 유지하면서 작은 객체에 대한 검출 성능을 향상시켰다. SSD를 기반으로 설계된 또다른 모델인 RefineDet \cite{zhang2018single}은 ARM(Anchor Refinement Module)을 활용하여 객체탐지를 진행한다. ARM은 객체 탐지 과정 중, 입력의 불필요한 배경을 없애서 분류기의 공간 탐색의 경우를 줄이게 한다.

대표적인 단일단계 객체탐지 모델인 YOLO는 입력 이미지를 특정 개수의 grid로 나누어 각각의 grid cell마다의 객체의 bounding box의 위치 및 해당 객체의 클래스를 추정한다. 이후 발전된 모델인 YOLOv3에서는 다양한 크기의 특징맵을 서로 다른 세 가지 다른 스케일로 추출하는 방법인 feature pyramid network를 적용해 mAP를 증가시켰다. YOLOv4 \cite{bochkovskiy2020yolov4}는 Backbone-Neck-Head로 구성되어 있으며, 기존 YOLOv3 모델의 Head 부분을 유지한 채, Backbone layer에 CSPDarknet53을 도입하고 Neck 부분에서 SPP(Spatial Pyramid Pooling), PAN(Path Aggregation Network) 구조를 적용하였다. 이와 더불어 학습 과정에서의 새로운 기법들을 도입하여 추론속도를 높이는 BoF(Bag of Freebies)와 약간의 추론속도를 희생하지만 더욱 정확도를 높이는 기법인 BoS(Bag of Specials)를 사용하여 전체적인 성능을 향상시켰다.

위에서 언급한 SSD와 YOLO 계열의 단일단계 방식의 검출기는 이 단계 방식의 검출기에 비해 상대적으로 빠른 속도로 추론이 가능하다. 실제로 대표적 단일단계 방식 검출기인 YOLOv4-s는 고성능 GPU인 NVIDIA GTX 1080 Ti 환경에서는 62.6 FPS와 같은 빠른 속도로 동작 가능하지만, 제한된 자원을 지닌 자율주행용 임베디드 플랫폼에서 실시간으로 구동되기에는 여전히 어려움이 존재한다.

\vspace{.25cm}
\subsection{객체탐지 모델 경량화 연구}
딥러닝 기반 객체탐지 모델을 저사양 환경에 탑재하기 위해 경량화된 객체탐지 모델 개발에 관한 연구들이 활발히 이뤄지고 있다. ResNet \cite{he2016deep}에서는 short-cut 기법을 사용한 residual block을 통해 잔 차를 줄여 나가는 방식으로 네트워크 깊이를 깊게 하여, 경량화를 이룸과 동시에 gradient vanishing 문제를 해결했다. SqueezeNet \cite{iandola2016squeezenet}에서는 fire module을 사용하여 경량화를 진행했다. Fire module의 squeeze layer에서 1x1 필터를 사용해 입력 채널의 수를 줄임으로 파라미터 수를 감소시켰고, expand layer에서 3x3 필터와 1x1 필터를 동시에 사용하여 50배의 모델 사이즈 경량화를 이뤘다.
Mobilenet \cite{howard2017mobilenets, sandler2018mobilenetv2}은 모바일 기기에서 실시간 객체 검출 알고리즘 탑재를 위해 개발된 모델로 DSC (Depthwise Separable Convolution) 기법을 사용한다. DSC 기법은 CNN에 사용되는 2d convolution을 각각의 채널에서 대해 convolution을 진행하는 depthwise convolution과 채널방향만을 고려하는 convolution인 pointwise convolution으로 분리하여 진행한다. 이를 통해, 파라미터 수와 연산량을 획기적으로 감소시켰다. ShuffleNet \cite{zhang2018shufflenet, ma2018shufflenet}은 입력 채널을 여러 그룹으로 나누어 convolution을 진행하는 group convolution 연산 과 각 group에 해당하는 채널들을 shuffle하는 channel shuffle 연산을 통해 모델 사이즈를 감소시킨다.

새로운 경량화 모델을 개발하는 것이 아니라, 기존 객체탐지 모델에 여러 경량화 기법들을 적용하는 연구들도 진행되고 있다. 대표적인 기법인 양자화(Quantization)는 기존 신경망의 가중치에 사용되는 비트 수를 감소시켜 연산량을 줄이는 방식이다. Jacob et al \cite{jacob2018quantization}은 ResNet, Inception 모델에 대해 기존의 32비트 float로 표현되는 가중치 정보를 8비트 integer로 표현함으로써 양자화를 진행하였고 4배의 모델 사이즈 감소를 이루었다. Yuan Cheng et al \cite{cheng2018deepeye}은 YOLOv3 모델에 양자화와 텐서화 기법을 적용하여 3.9배의 모델 압축을 이뤄냈다. 

하지만, 위에 언급된 논문에서 제안된 Q-YOLO와 T-RNN은 NVIDIA GTX 1060 Ti 기반 GPU에서 24~25 FPS의 값을 가지므로 자율주행자동차를 위한 임베디드 GPU 플랫폼에서 실시간 운용하기에는 한계가 있음을 알 수 있다. 고로, 본 연구에서는 임베디드 GPU 환경에서 실시간 구동이 가능한 경량화 객체탐지 모델인 YOffleNet을 제안하고자 한다.

\vspace{.25cm}
\section{객체탐지 경량화 알고리즘 설계}
\vspace{.25cm}
\subsection{Background}

\vspace{.15cm}
\subsubsection{YOLOv4}
대표적인 단일단계 방식의 검출기인 YOLO는 이미지로부터 한 번에 객체의 class와 bounding box를 예측하는 방식으로 이 단계 방식의 검출기보다 빠른 추론이 가능하다는 장점을 지닌다. YOLOv2는 기존 YOLOv1 모델의 모든 convolution layer에 batch normalization을 적용함으로써 regularization 효과를 얻어 정확도를 높였고, YOLOv3는 각 class에 대해 독립적으로 로지스틱 회귀를 적용하였고 3개의 다른 크기의 feature에 대해 각각 prediction을 진행하는 FPN을 도입하여 정확도를 향상시켰다. YOLOv4는 Backbone layer에 CSP DenseNet을 적용한 CSPDarknet53을 도입하고 Neck 부분에 SPP와 PAN, FPN(Feature Pyramid Network) 구조를 도입하여 단일 GPU 환경에서도 높은 정확도와 빠른 객체탐지가 가능하도록 설계되었으며, 기존 YOLOv3에 비해 mAP와 FPS가 각각 10\%, 12 정도 향상된 성능을 보여준다. 따라서, 본 연구에서는 실시간 구동을 위해 가장 많이 사용되는 모델인 YOLOv4를 객체탐지를 위한 경량화 모델의 baseline 모델로 선정하였다.

YOLOv4의 Backbone layer인 CSPDarknet53에서는 입력 데이터를 특정 횟수만큼 convolution 한 후 복사하여 이전 layer와 concatenation하는 기법을 활용하는 CSP DenseNet \cite{wang2020cspnet} 모듈을 활용한다. 이를 통해 이전 피처의 가중치를 재사용하고 중복되는 gradient 값을 줄여 모델을 경량화한다.

Neck에서는 입력 특징맵에 대해 3가지 커널 사이즈로 max pooling한 후 concatenation을 진행하는 SPP \cite{he2015spatial} 기법을 통해 여러 크기의 입력에 대한 정확도를 향상시켰다. 또한 Bottom-up path augmentation 기법을 통해 원활한 정보흐름을 가능하게 하는 PANet \cite{liu2018path} 구조를 적용시켜 정확도를 높였다. PANet은 Convolution layer로 이뤄진 피라미드 구조에 최하단 특징맵과의 Short-cut과 Convolution layer로 구성된 새로운 피라미드 구조를 만든다. YOLOv4에서는 PANet을 도입하여 정보의 흐름을 원활하게 하며 경량화를 통해 발생되는 정확성 손실의 문제를 보완한다.

Head의 경우 YOLOv3의 객체탐지 방식을 그대로 도입하였다. FPN(Feature Pyramid Network)에 의해 생성된 3개의 특징맵에 대한 Detection이 진행되며 Anchor box 사이즈의 경우 훈련 이미지를 k-means clustering으로 분석하여 결정된다. Head의 탐지과정은 각각의 bounding box마다의 box가 객체를 포함하고 있을 가능성을 예측하고 이때 GT(Ground Truth)와 IOU(Intersection over Union)가 가장 높은 bounding box를 1로 설정해 GT마다 1개의 bounding box를 할당한다. Classification 부분에서는 softmax 대신 로지스틱 회귀를 적용시켜 multi label classification을 가능하게 한다.

\begin{figure}[t]
    \captionsetup{singlelinecheck = false, format= hang, justification=raggedright, labelsep=space}
    \centering
        \includegraphics[width=\linewidth]{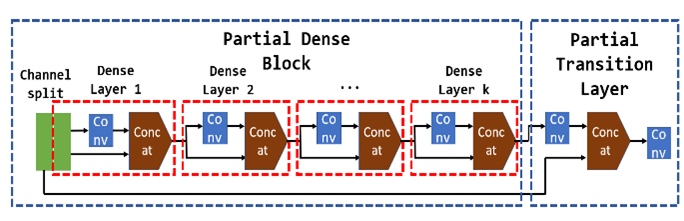}
    \caption{CSP DenseNet mdoule}
    \label{CSPDenseNet}
\end{figure}

\vspace{.15cm}
\subsubsection{ShuffleNet v2 모듈}
본 연구에서 사용되는 ShuffleNet v2 \cite{ma2018shufflenet} 모듈은 channel split을 통해 파라미터 수와 연산량을 감소시키며, channel shuffle 연산을 통해 정보의 흐름을 원활히 하는 효과를 얻는다. ShuffleNet v2는 파라미터 수를 줄이는 것뿐만 아니라, MAC(Memory Access Cost)를 줄여 딥러닝 모델의 추론 속도를 높이는 방법을 제시한다.

ShuffleNet v2 모듈의 경우 channel split을 통해 입력 특징맵의 채널을 절반으로 줄인 뒤 convolution 연산을 진행하여 연산량을 절반으로 감소시켰다. 나머지 절반의 입력 피처맵에 대해서는 convolution 연산을 거친 특징맵과 short-cut 구조를 통해 concatenation함으로써 정보손실 방지에 유리한 구조를 가진다.

ShuffleNet v2에서 MAC를 최소화하기 위해 여러가지 전략들을 도입했다. 1x1 convolution 시 입출력 특징맵의 채널 크기를 동등하게 가져가는 것이 유리함을 입증하여 동등한 입출력의 채널 크기를 가지는 1x1 convolution을 도입한다. 또한, group convolution 연산 과정에서 group의 크기가 증가하면 MAC가 증가한다는 것을 증명한다. 마지막으로, Element-wise addition과 같은 Element-wise operation이 MAC를 크게 증가시켜 결과적으로 런타임을 크게 증가시킨다는 사실을 언급한다. 이러한 사실을 바탕으로 channel split된 특징맵을 합치는 과정에서 elementwise addition 대신 concatenation 연산으로 대체하여 런타임을 감소시킨다.

\begin{figure}[t]
    \captionsetup{singlelinecheck = false, format= hang, justification=raggedright, labelsep=space}
    \centering
        \includegraphics[width=\linewidth]{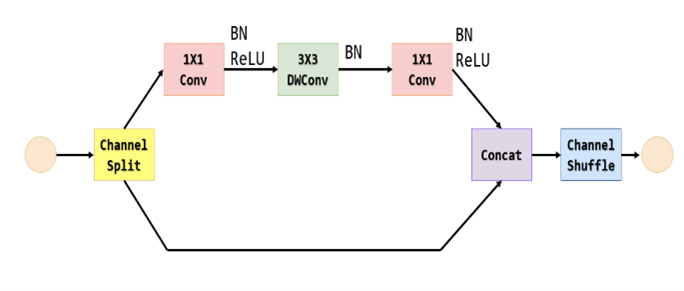}
    \caption{ShuffleNet v2 module}
    \label{ShuffleNetv2Module}
\end{figure}

\vspace{.25cm}
\subsection{YOLOv4 경량화 설계 전략}
YOLOv4에서 사용된 Fig \ref{ShuffleNetv2Module}의 CSP DenseNet는 convolution과 concatenation이 반복될수록 파라미터 수와 연산량이 증가한다. 또한, 초기 특징맵의 정보 손실을 방지하기 위해 PANet 구조는 Bottom-up path augmentation을 통해 정보의 흐름을 원활하게 하지만 피처맵의 채널의 수를 증가시켜 파라미터 수를 증가시키고 이는 연산량을 증가시킨다. 본 연구에서 제안하는 YOffleNet 모델은 위에서 언급한 YOLOv4 모델의 단점을 개선하고 경량화 효과를 이루기 위해 아래의 방법들을 제시한다.

\begin{figure*}[t]
    \captionsetup{singlelinecheck = false, format= hang, justification=raggedright, labelsep=space}
    \centering
        \includegraphics[width=\linewidth]{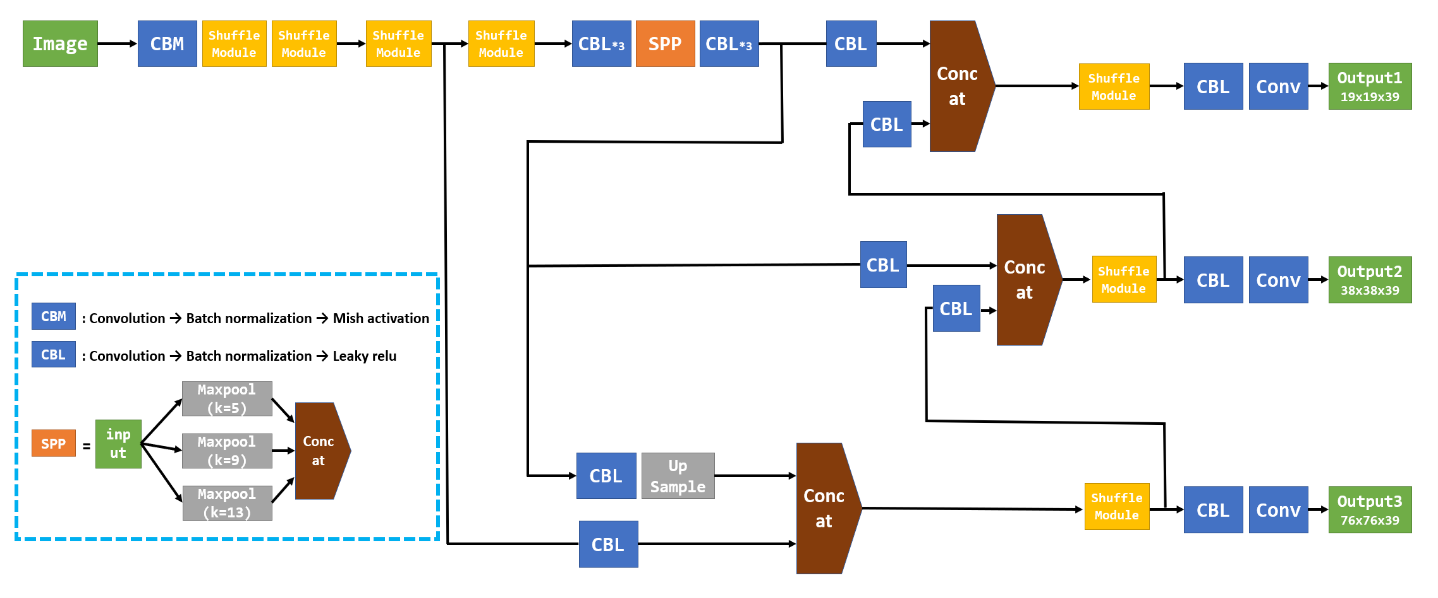}
    \caption{Overall Architecture of YOffleNet}
    \label{YOffleNetArchitecture}
\end{figure*}

\begin{figure}[t]
    \captionsetup{singlelinecheck = false, format= hang, justification=raggedright, labelsep=space}
    \centering
        \includegraphics[width=\linewidth]{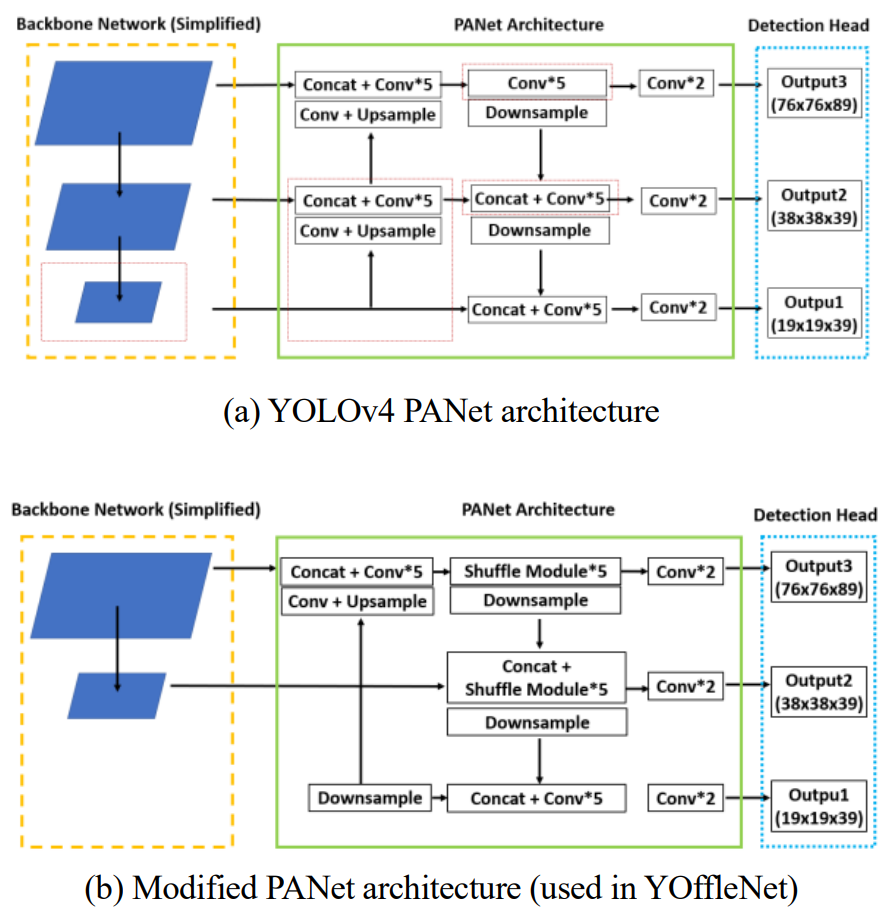}
    \caption{PANet architecture in YOLOv4 and YOffleNet}
    \label{PANetSimplification}
\end{figure}

\noindent1) Backbone layer인 CSP DenseNet(Fig \ref{CSPDenseNet})는 네트워크가 깊어짐에 따라 연산량이 필연적으로 증가하는 구조이다. 본 연구에서는 경량화된 모델 설계를 위해 backbone layer를 ShuffleNet (Fig \ref{ShuffleNetv2Module}) 모듈로 구성했다.

\noindent2) YOLOv4 네트워크에서 사용하는 SPP + PANet 구조를 간소화하여 모델의 크기를 경량화 시킨다. 기존 YOLOv4 모델의 PANet은 backbone network로부터 3개의 층으로 구성된 입력을 받도록 설계되어 있다. 하지만, 다양한 크기 및 종류의 클래스가 존재하는 일반적인 객체검출 상황과 달리 자율주행 환경에서는 제한된 클래스에 객체검출(자동차, 보행자 등)을 진행하게 된다. 

이러한 점을 고려하여, 변경된 PANet에서는 backbone network로부터 2개의 층으로만 구성된 입력을 받으며, 그에 따라 Upsample, Downsample layer의 위치 및 개수를 변경하였다. 위의 두 기법을 통해 제안하는 YOffleNet 모델의 전체적인 구조는 Fig \ref{YOffleNetArchitecture}와 같다.

\vspace{.25cm}
\section{실험 및 검증}
\vspace{.25cm}
\subsection{실험환경 구성}
제안하는 YOffleNet 모델의 학습 및 성능평가를 위해 라벨링된 KITTI data 7480장의 이미지 데이터셋을 확보했다. 확보된 데이터는 Table \ref{NumberofDataset}과 같이 학습, 검증 및 시험용 데이터셋으로 분류하여 실험을 진행했다. 시험용 데이터는 학습에는 전혀 사용되지 않고 학습 후 모델 평가를 위해 구분해 두었다.

\begin{table}[h]
\captionsetup{singlelinecheck = false, format= hang, justification=raggedright, labelsep=space}
\begin{center}
\begin{tabular}{|c|c|c|c|}
\hline
Data & Training & Validation & Test \\
\hline
Counts & 3740 & 2244 & 1496 \\
\hline
\end{tabular}
\end{center}
\caption{Number of training, validation and test KITTI dataset}
\label{NumberofDataset}
\end{table}

학습 과정에서 16의 batch size로 300 epoch 동안 훈련을 진행했으며, 과적합(overfitting)을 방지하기 위해 batch normalization을 적용했다. NVIDIA GTX 1080 Ti GUI 환경에서 약 10시간 정도 학습을 진행했으며, 학습을 마친 후 GIOU Loss(Generalized Intersection over Union Loss) 값이 약 0.027에 수렴하여 충분히 학습이 이루어진 것을 확인했다.

\vspace{.25cm}
\subsection{성능평가}
YOffleNet의 성능평가를 위한 비교모델로서 대표적인 단일단계 방식의 객체탐지 모델인 YOLOv4-s와 YOLOv4-m, YOLOv3, 대표적 경량화 객체탐지 모델인 YOLOv3-mobilenet, YOLOv3 tiny을 선정하여 성능비교를 진행했다. YOLOv4의 m 버전은 s 버전보다 모델의 크기가 크지만 정확도는 더 높고, s 버전은 m 버전보다 모델의 크기가 작지만 정확도는 더 낮다. YOLOv4는 YOffleNet(ours)의 기반이 되기 때문에 YOLOv4 s, m 모델과 정확도 및 속도 성능 비교를 진행했다.

또한, YOLOv4의 이전 모델인 YOLOv3와 YOLOv3의 경량화 버전인 YOLOv3-tiny도 평가모델로 선정하였다. YOLOv3-mobilnet은 YOLOv3 모델의 backbone을 Mobilenetv332)로 대체한 모델이며, Mobilenet의 경우 경량화 모델에 널리 사용되기 때문에 비교모델로서 적합하다 판단했다.

학습 완료 후 학습된 모델의 성능 평가를 위해 시험용 데이터셋으로부터 객체 검출 성능 테스트를 수행하였다. 시험용 데이터셋은 학습에 사용된 이미지 외의 1658장의 이미지를 사용하였다. 성능 검증을 위해 모델의 가중치 크기, mAP, 파라미터 수, FPS를 검증 지표로 사용했다.

\begin{table}[t]
\tabcolsep=0.11cm

\begin{tabular}{c|c|c|c|c}  \hline 
                                                            & mAP    & \begin{tabular}[c]{@{}c@{}}Weight\\ file size\end{tabular} & \begin{tabular}[c]{@{}c@{}}\# of\\ parameters\end{tabular} & \begin{tabular}[c]{@{}c@{}}FPS\\ (GTX\\ 1080 Ti)\end{tabular} \\ \hline \hline
\multicolumn{5}{l}{ \textbf{\textit{Deep CNN Model:}}}                                                                                                                                                                                                                            \\ \hline
\begin{tabular}[c]{@{}c@{}}YOLOv4-s\\ (base)\end{tabular}   & 88.4\% & 18.5 MB                                                    & 9.1 M                                                      & 138.9                                                         \\ \hline
YOLOv4-m                                                    & 89.4\% & 49.1 MB                                                    & 24.4 M                                                     & 107.5                                                         \\ \hline
YOLOv3                                                      & 90.5\% & 123.4 MB                                                   & 61.5 M                                                     & 76.3                                                          \\ \hline \hline
\multicolumn{5}{l}{ \textbf{\textit{Compressed Model:}}}                                                                                                                                                                                                                          \\ \hline
YOLOv3 tiny                                                 & 81.1\% & 17.4 MB                                                    & 8.7 M                                                      & 357.1                                                         \\ \hline
\begin{tabular}[c]{@{}c@{}}YOLOv3-\\ mobilenet\end{tabular} & 70.2\% & 95.8 MB                                                    & 23.8 M                                                     & 98.0                                                          \\ \hline \hline
\multicolumn{5}{l}{ \textbf{\textit{Ours:}}}                                                                                                                                                                                                                                      \\ \hline
YOffleNet (S)                                               & 87.5\% & 17.1 MB                                                    & 8.5 M                                                      & 161.3                                                         \\ \hline
YOffleNet (P)                                               & 88.2\% & 14.9 MB                                                    & 7.3 M                                                      & 137.0                                                         \\ \hline
\begin{tabular}[c]{@{}c@{}}YOffleNet\\ (S+P)\end{tabular}   & 85.8\% & 4.0 MB                                                     & 1.9 M                                                      & 196.1                                                      \\   \hline
\end{tabular}
\caption{Performance results of each model in KITTI Dataset \\
(S): applying ShuffleNet module \\
(P): applying PANet simplification}
\label{PerformanceTable1}
\end{table}

Table \ref{PerformanceTable1}에서 weight file size는 학습 후 도출되는 모델의 크기를 의미하며 YOffleNet의 weight file size는 YOLO v4-m에 비해 약 12배 작고 YOLO v4-s에 비해 약 4.6배 작음을 확인할 수 있다. 이는 본 연구의 모델이 평가모델에 비해 간단한 구조를 가지고 있음을 의미하며 임베디드 플랫폼에서 실시간 구동에 이점을 갖는 것을 확인할 수 있다. mAP의 경우 가장 정확도가 높은 비교모델인 YOLO v3에 비해 4.7\% 정도의 차이를 보였다. 경량화 객체탐지 모델인 YOLOv3 tiny, YOLOv3-mobilenet과 비교했을 때도 모델 사이즈 측면에서 4.4배, 23.9배 경량화 된 것을 확인했다.

추론 속도 성능인 FPS의 경우 Table \ref{PerformanceTable2}에서 확인할 수 있듯 CPU 환경에서는 YOLov3-tiny가 28.6 FPS로 비교모델 중 가장 빠르게 동작했지만, Jetson Nano 보드에서는 제안하는 YOffleNet 모델이 약 9.9 FPS로 가장 빠른 추론 속도를 가지는 것이 확인됐다. 이는 GPU 상에서 MAC를 감소시키기 위한 전략을 취한 ShuffleNet 모듈을 적용한 효과이며, YOffleNet의 base모델인 YOLO v4-s 와 비교했을 때 약 2.36배 빠른 FPS로 동작하는 확인할 수 있다.

\begin{table}[t]
\tabcolsep=0.11cm
\begin{tabular}{c|c|c|c|c}  \hline
                                                            & \begin{tabular}[c]{@{}c@{}}PC GPU\\ (GTX\\ 1080Ti)\end{tabular} & \begin{tabular}[c]{@{}c@{}}Jetson\\ AGX\\ Xavier\end{tabular} & \begin{tabular}[c]{@{}c@{}}Jetson\\ Nano\end{tabular} & PC CPU \\ \hline \hline
\multicolumn{5}{l}{ \textbf{\textit{Deep CNN Model:}}}                                                                                                                                                                                                                            \\ \hline
\begin{tabular}[c]{@{}c@{}}YOLOv4-s\\ (base)\end{tabular}   & 138.9                                                           & 33.8                                                          & 4.2                                                   & 20.8   \\ \hline
YOLOv4-m                                                    & 107.5                                                           & 31.1                                                          & 2.3                                                   & 11.8   \\ \hline
YOLOv3                                                      & 76.3                                                            & 30.6                                                          & 1.6                                                   & 6.8    \\ \hline \hline
\multicolumn{5}{l}{ \textbf{\textit{Compressed Model:}}}                                                                                                                                                                                                                          \\  \hline 
YOLOv3 tiny                                                 & 357.1                                                           & 51.5                                                          & 6.5                                                   & 28.6   \\ \hline
\begin{tabular}[c]{@{}c@{}}YOLOv3-\\ mobilenet\end{tabular} & 98.0                                                            & 22.1                                                          & 4.5                                                   & 17.9   \\ \hline \hline
\multicolumn{5}{l}{ \textbf{\textit{Ours:}}}                                                                                                                                                                                                                                      \\ \hline
YOffleNet (S)                                               & 161.3                                                           & 42.0                                                          & 7.6                                                   & 24.1   \\ \hline
YOffleNet (P)                                               & 137.0                                                           & 35.3                                                          & 5.9                                                   & 22.7   \\ \hline
\begin{tabular}[c]{@{}c@{}}YOffleNet\\ (S+P)\end{tabular}   & 196.1                                                           & 46.1                                                          & 9.9                                                   & 21.6  \\ \hline
\end{tabular}
\caption{FPS of each model in several environment \\
(S): applying ShuffleNet module \\
(P): applying PANet simplification}
\label{PerformanceTable2}
\end{table}

\vspace{.25cm}
\section{결론}
본 연구에서는 자율주행용 임베디드 플랫폼 환경에 탑재가능한 딥러닝 객체탐지 모델인 YOffleNet을 제안했고, KITTI 데이터셋을 통해 제안하는 모델의 성능 검증을 진행했다. YOffleNet은 대표적인 단일단계 방식 검출기인 YOLOv4를 기반으로 하며, 연산량 감소를 위해 PANet을 구조적으로 간소하고 Backbone layer에 사용되는 CSP DenseNet 모듈을 ShuffleNet 모듈로 대체하는 구조 경량화를 진행했다.

대표적인 자율주행용 데이터셋인 KITTI 데이터셋을 사용해 학습을 진행했으며, 본 논문에서 기반으로 삼은 모델인 YOLOv4와 비교했을 때 YOLOv4-s에 비해 mAP는 2.6\% 감소했지만, 4.6배 감소된 weight file 크기를 가지는 것이 확인됐다. 임베디드 플랫폼 상에서의 실시간 구동을 확인하기 위해 YOffleNet 모델을 NVIDIA Jetson AGX Xavier 보드에 탑재한 결과 기존 YOLOv4에 비해 12.3 FPS 증가한 46.1 FPS 로 동작함을 확인했다. 추후 연구에서는 YOffleNet 모델에 Quantization, Network Pruning, Distillation 등의 경량화 기법을 추가하여 Jetson Nano와 Raspberry pi 4B 와 같은 보다 저사양의 임베디드 플랫폼에서의 실시간 구동 가능성을 확인할 것이다.

\begin{figure*}[t]
    \captionsetup{singlelinecheck = false, format= hang, justification=raggedright, labelsep=space}
    \centering
        \includegraphics[width=\linewidth]{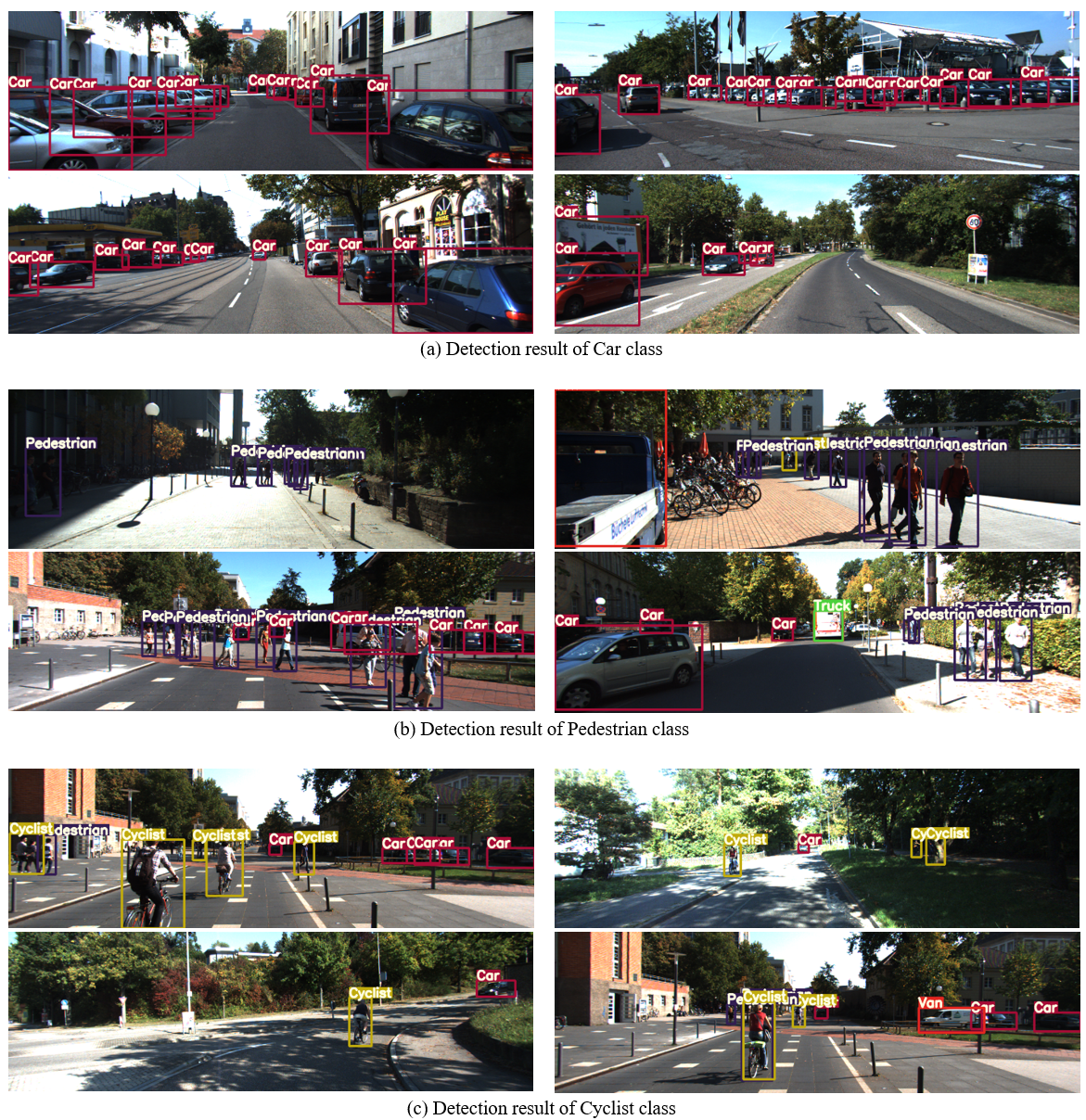}
    \caption{Detection result of KITTI images}
    \label{Results}
\end{figure*}

\pagebreak
{\small
\bibliographystyle{ieee}
\bibliography{egbib}
}

\end{document}